\def\year{2020}\relax
\documentclass[letterpaper]{article} 
\usepackage{aaai20}  
\usepackage{times}  
\usepackage{helvet} 
\usepackage{courier}  
\usepackage[hyphens]{url}  
\usepackage{graphicx} 
\usepackage{graphicx}  
\usepackage{amsfonts}
\usepackage{amsmath}
\usepackage{amsthm}
\usepackage{multirow}
\usepackage[para]{threeparttable}
\usepackage{booktabs}
\DeclareMathOperator{\AGRU}{\mathbf{AGRU}}
\DeclareMathOperator{\GRU}{\mathbf{GRU}}
\DeclareMathOperator{\maxpool}{maxpool}
\DeclareMathOperator{\relu}{relu}
\DeclareMathOperator{\softmax}{softmax}

\title{Real-Time Emotion Recognition via Attention Gated \\Hierarchical Memory Network}
\author{
Wenxiang Jiao,\textsuperscript{\rm 1}
Michael R. Lyu,\textsuperscript{\rm 1}
Irwin King\textsuperscript{\rm 1}\\
\textsuperscript{\rm 1}Department of Computer Science and Engineering,\\
The Chinese University of Hong Kong, Shatin, N.T., Hong Kong, China \\
\{wxjiao, lyu, king\}@cse.cuhk.edu.hk
}

\begin{document}

\maketitle

\begin{abstract}
Real-time emotion recognition (RTER) in conversations is significant for developing emotionally intelligent chatting machines. Without the future context in RTER, it becomes critical to build the memory bank carefully for capturing historical context and summarize the memories appropriately to retrieve relevant information.
We propose an Attention Gated Hierarchical Memory Network (AGHMN) to address the problems of prior work: (1) Commonly used convolutional neural networks (CNNs) for utterance feature extraction are less compatible in the memory modules; (2) Unidirectional gated recurrent units (GRUs) only allow each historical utterance to have context before it, preventing information propagation in the opposite direction; (3) The Soft Attention for summarizing loses the positional and ordering information of memories, regardless of how the memory bank is built.
Particularly, we propose a Hierarchical Memory Network (HMN) with a bidirectional GRU (BiGRU) as the utterance reader and a BiGRU fusion layer for the interaction between historical utterances. For memory summarizing, we propose an Attention GRU (AGRU) where we utilize the attention weights to update the internal state of GRU. We further promote the AGRU to a bidirectional variant (BiAGRU) to balance the contextual information from recent memories and that from distant memories. We conduct experiments on two emotion conversation datasets with extensive analysis, demonstrating the efficacy of our AGHMN models.
\end{abstract}

\section{Introduction}

\begin{figure}[t!]
    \centering
    \includegraphics[width=0.95\columnwidth]{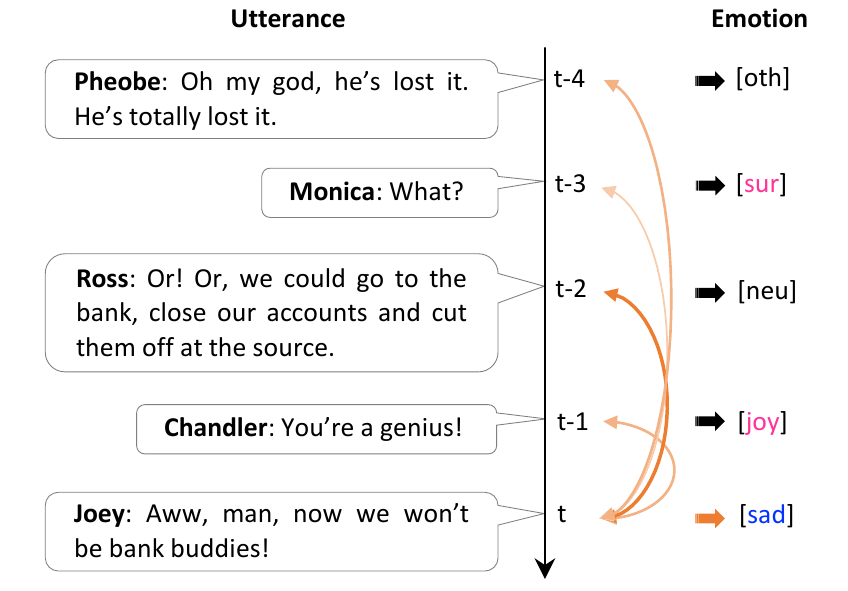}
    \caption{The specification of the RTER task.}
    \label{fig:task_spec}
\end{figure}

Emotion recognition is a significant research topic because of the potential application in developing empathetic machines in present artificial intelligence (AI) era.
We focus on the real-time scenario to detect the emotion state of speakers in an ongoing conversation at utterance-level. According to \cite{olson1977utterance}, an utterance is a unit of speech bounded by breathes and pauses. We term this task as Real-Time Emotion Recognition (RTER). Inherently, emotion recognition is a multi-modal learning task that could involve text, video and audio features, but the text feature plays the most significant role~\cite{DBLP:conf/fgr/ChenHMN98,DBLP:conf/emnlp/PoriaCG15,DBLP:conf/acl/PoriaCHMZM17,DBLP:conf/naacl/HazarikaPZCMZ18}. Thus, in this paper, we tackle the RTER task in text conversations. 

Without the future context, in RTER, it becomes critical to exploit the contextual information from the historical utterances. For this purpose, one needs to take good care of two factors, i.e., the memory bank for capturing historical context, and the summarizing technique for the query to extract relevant information from the memory bank.
The memory bank is usually built in a two-level fashion to simulate the hierarchical structure of conversations, i.e., words-to-utterance and utterances-to-conversation. Specifically, existing models~\cite{DBLP:conf/emnlp/HazarikaPMCZ18,DBLP:conf/naacl/HazarikaPZCMZ18,DBLP:conf/aaai/MajumderPHMGC19} obtain their memory banks by utilizing convolutional neural networks~(CNNs) to learn utterance features and unidirectional gated recurrent units~(GRUs)~\cite{DBLP:conf/emnlp/ChoMGBBSB14} to capture relationship of utterances. However, through our exploration, we find that a bidirectional GRU~(BiGRU) learns better utterance features than commonly used CNNs. Moreover, the unidirectional GRU only allows each historical utterance to have context before but not after it, which may prevent information propagation in the opposite direction. As for the summarizing techniques, the commonly used Soft Attention produces a weighted sum of the memories, which can be regarded as a bag-of-memories. Just as the bag-of-words in word representation area~\cite{DBLP:conf/nips/MikolovSCCD13} that lacks sensitivity to word order~\cite{DBLP:conf/emnlp/LingTAFDBTL15}, the bag-of-memories loses the positional and ordering information of the memories, regardless of how the memory bank is built.

Incorporating these factors, in this paper, we propose an Attention Gated Hierarchical Memory Network (AGHMN) to better extract the utterance features and the contextual information for the RTER task. Specifically, we summarize our contributions as below: (1) We propose a Hierarchical Memory Network (HMN) to improve the utterance features and the memory bank for extracting contextual information. The HMN is essentially a two-level GRU encoder, including an utterance reader and a fusion layer. 
The utterance reader applies a BiGRU to model the word sequence of each utterance, which we demonstrate that it is more compatible with the hierarchical structure. The fusion layer adopts a BiGRU to read the historical utterances, which enables sufficient interaction between them. 
(2) We propose an Attention GRU (AGRU) to retain the positional and ordering information while summarizing the memories, and promote it to its bidirectional variant, i.e., BiAGRU, to capture more comprehensive context. The AGRU is formed by utilizing the attention weights of the query to the memories to update the internal state of a normal GRU. The final hidden state of AGRU serves as the contextual vector to help refine the representation of the query. The BiAGRU is dedicated to balancing the information from recent memories and that from distant memories.
(3) We conduct experiments on two emotion conversation datasets with extensive analysis, demonstrating the efficacy of our proposed AGHMN models.

\section{Related Work}

\noindent\textbf{Text Classification.}
Text-based emotion recognition is usually treated as a text classification problem. Previously proposed methods can be mainly divided into three categories: keyword-based methods~\cite{DBLP:conf/aaai/WilsonWH04}, learning-based methods~\cite{DBLP:conf/webi/YangLC07}, and hybrid methods~\cite{DBLP:journals/talip/WuCL06}. 
Nowadays, deep learning is dominating the text classification area due to its powerful capability in learning latent features. Representative methods include convolutional neural network (CNN)~\cite{DBLP:conf/emnlp/Kim14}, recurrent neural network (RNN)~\cite{DBLP:conf/acl/Abdul-MageedU17}, and hierarchical attention network (HAN)~\cite{DBLP:conf/emnlp/TangQL15}. These work are customized for data unit without context, e.g. independent reviews or documents.
\\[1ex]
\noindent\textbf{Context-Dependent Models.}
Recognizing emotion state of speakers in conversations requires that the query should take into account the context to carry accurate information. Existing work can be divided into two streams: the static models, and the dynamic models. The static models include sequence-based and graph-based~\cite{DBLP:conf/ijcai/ZhangWSLZZ19,DBLP:journals/corr/abs-1908-11540}, the former of which let each utterance to have the utterances both in the history and the future as context. Among sequence-based static models, cLSTM~\cite{DBLP:conf/acl/PoriaCHMZM17} only adopts long short-term memory networks~(LSTMs)~\cite{DBLP:journals/neco/HochreiterS97} to capture the sequential relationship between the utterances. HiGRU~\cite{DBLP:conf/naacl/JiaoYKL19} employs a self-attention mechanism for context weighting and summarizing, as well as a residual connection for feature fusion. BiDialogueRNN ~\cite{DBLP:conf/aaai/MajumderPHMGC19} is built on RNNs that keeps track of the individual party states throughout the conversation and uses this information for emotion recognition. These static models may adapt to the RTER task if we take their unidirectional variants. The dynamic models read the utterances in the order as they are generated so that each incoming utterance, i.e., the query, only depends on the historical utterances. These models include CMN~\cite{DBLP:conf/naacl/HazarikaPZCMZ18}, DialogueRNN, and ICON~\cite{DBLP:conf/emnlp/HazarikaPMCZ18}. Among them, CMN and ICON are customized for dyadic conversations, which incorporate memory networks~\cite{DBLP:conf/nips/SukhbaatarSWF15} to refine the contextual information and also consider the self- and inter-speaker emotional influence. 

Our AGHMN models differ from these approaches in that we aim to produce better utterance features and memory representation by our proposed HMN and summarize the memories in a better way by our proposed AGRU and BiAGRU. We do not distinguish speakers explicitly as in DialogueRNN but find that the model itself can recognize the difference between speakers (see \textbf{Case Study}).

\section{Task Specification}

We first specify the task of Real-Time Emotion Recognition (RTER) as below:
\\[1ex]
\noindent\textbf{Real-Time Emotion Recognition.}
Suppose that a conversation has proceeded for $t$ turns so far with the utterance sequence $\mathcal{C}_t = \{u_1, \cdots, u_t\}$, the $t$-th utterance is the query utterance $q$, and the others are historical ones. 
As illustrated in Fig~\ref{fig:task_spec}, each utterance expresses a major emotion $e$ among a set of emotions $\mathcal{E}$, such as joy, sadness, and neutral. Our goal is to design and train a model $\mathcal{M}$ to predict the emotion expressed by $q$ conditioned on the historical utterances.

\begin{figure*}[!t]
    \centering
    \includegraphics[width=0.95\textwidth]{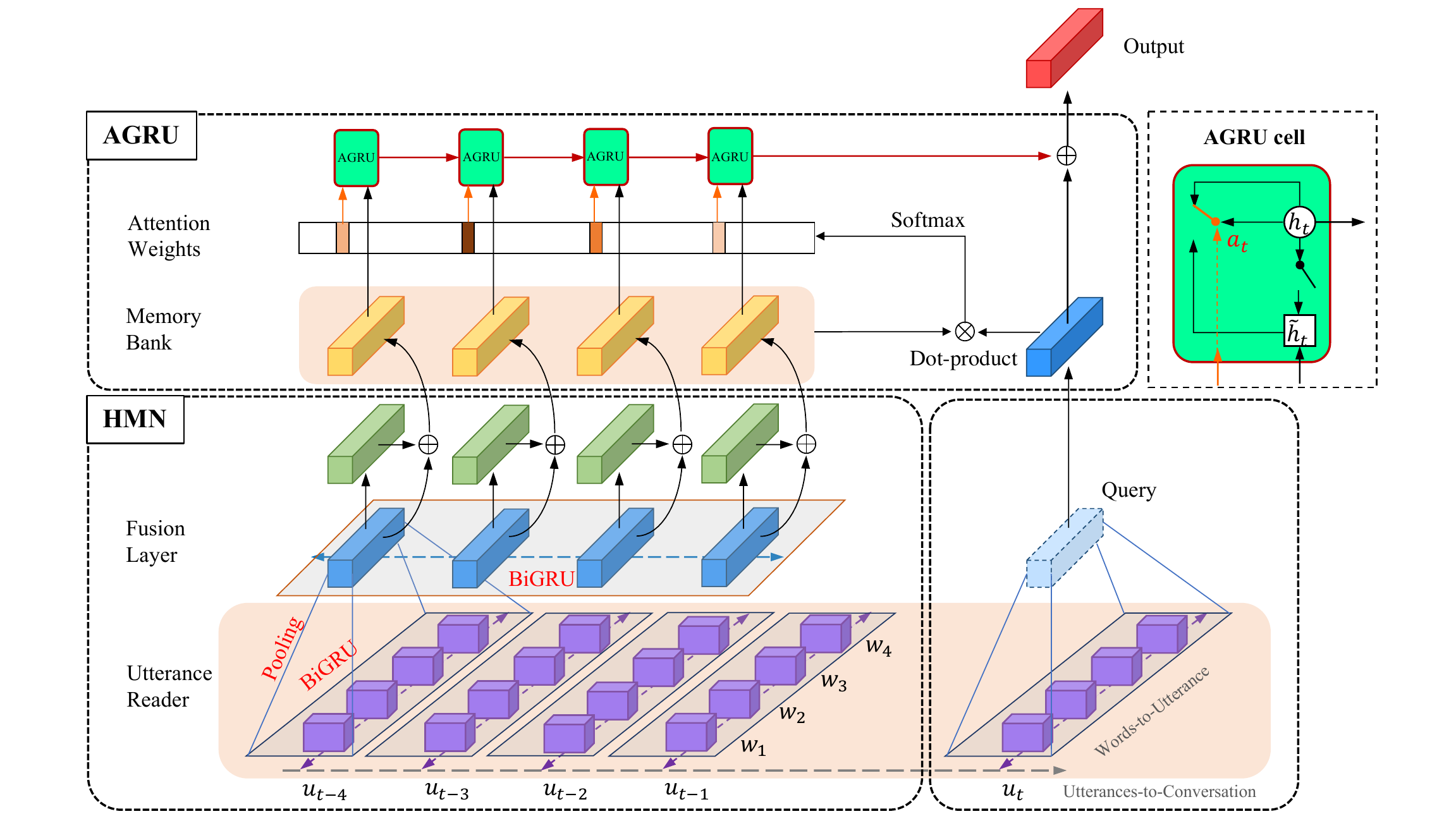}
    \caption{The architecture of our AGHMN model.}
    \label{fig:AGHMN}
\end{figure*}

\section{Architecture}

In this section, we will introduce the AGHMN model as illustrated in Fig~\ref{fig:AGHMN}, which consists of a Word Embedding Layer, a Hierarchical Memory Network, an Attention GRU, and a Classifier. 

\subsection{Word Embedding Layer}
For an utterance in $\mathcal{C}_t$ composed by a sequence of words $u_{t'}=\{w_1, w_2, \cdots, w_N\}$, where $t'\in [1, t]$, $N$ is the length of the utterance, and $w_n\in u_{t'}$ is the index of the word in the vocabulary. The utterance is fed into the word embedding layer to get a dense vector $\mathbf{x}\in\mathbb{R}^{d_w}$ for each word, where $d_w$ is the size of the word vector. The weights of the word embedding layer are initialized by the publicly available 300-dimensional word2vec~\cite{DBLP:conf/nips/MikolovSCCD13} vectors\footnote{\url{https://code.google.com/archive/p/word2vec/}} trained on 100 billion words from Google News. The words not included in the word2vec vocabulary will be initialized by randomly generated vectors.

\subsection{Hierarchical Memory Network}
The Hierarchical Memory Network (HMN) is a two-level encoder, the lower one as an utterance reader and the upper one as the fusion layer. 
\\[1ex]
\noindent\textbf{Utterance Reader.}
Though current work~\cite{DBLP:conf/emnlp/HazarikaPMCZ18,DBLP:conf/naacl/HazarikaPZCMZ18,DBLP:conf/aaai/MajumderPHMGC19} always utilize CNN to extract utterance features, we decide to adopt a BiGRU. The BiGRU is able to model the word sequence while gather the contextual information for each word in two directions, making it better for understanding a sentence sufficiently. Actually, we find that the BiGRU performs much better than a commonly used 1-D CNN as the utterance reader (see Table~\ref{table:result_ReaderCompare}). Specifically, the BiGRU takes an utterance represented by word vectors $\mathbf{X}=\{\mathbf{x}_n\}_{n=1}^{N}$ as input:
\begin{align}
    \overrightarrow{\mathbf{h}}_{n} &= \overrightarrow{\GRU}(\mathbf{x}_n, \overrightarrow{\mathbf{h}}_{n-1}), \\
    \overleftarrow{\mathbf{h}}_{n} &= \overleftarrow{\GRU}(\mathbf{x}_n, \overleftarrow{\mathbf{h}}_{n+1}),
\end{align}
where $\overrightarrow{\mathbf{h}}_{n},\overleftarrow{\mathbf{h}}_{n}\in\mathbb{R}^{d_1}$ are the hidden states of the forward GRU and the backward one, respectively, and $d_1$ is the hidden size. The hidden states of both directions are concatenated and fed to the max-over-time pooling layer. The resulted vector $\overleftrightarrow{\mathbf{h}}$ is then transformed to be the utterance embedding $\mathbf{u}\in \mathbb{R}^{d_1}$ through a $\tanh$ layer:
\begin{align}
\overleftrightarrow{\mathbf{h}} &= \maxpool(\{[\overrightarrow{\mathbf{h}}_{n},\overleftarrow{\mathbf{h}}_{n}]\}_{n=1}^N) \\
\mathbf{u} &= \tanh(\mathbf{W}_u\overleftrightarrow{\mathbf{h}} + \mathbf{b}_u),
\end{align}
where $\mathbf{W}_u\in \mathbb{R}^{d_1\times 2d_1}$, and $\mathbf{b}_u\in \mathbb{R}^{d_1}$.
\\[1ex]
\noindent\textbf{Fusion Layer.}
At $t$-step, the representation of the query comes from the utterance encoder: $\mathbf{q}_t = \mathbf{u}_t$.
For each query $\mathbf{q}_t$, we build a memory bank $\mathbf{M}_t$ based on the most recent $K$ historical utterances. Since the $K$ utterances maintain a sequence, we hope to make them interact with each other so as to refine the memory representation. As shown in Fig~\ref{fig:AGHMN}, we consider two types of memory banks here:

$\bullet$~\textbf{Unidirectional Fusion (UniF).} Firstly, we utilize a unidirectional GRU to read these $K$ utterances to model the sequential relationship between them. The independent utterance embeddings $\{\mathbf{u}_{t-K-1+k}\}_{k=1}^{K}$ are fed to the GRU, and are then connected to the output of the GRU to form the memory bank: $\mathbf{M}_t = \{\overrightarrow{\GRU}(\mathbf{u}_{t-K-1+k}) + \mathbf{u}_{t-K-1+k}\}_{k=1}^{K}$.

$\bullet$~\textbf{Bidirectional Fusion (BiF).} The UniF memory bank only allows each memory to have the context before it but not after it, which may prevent information propagation from the opposite direction. To address such a problem, we propose to read the $K$ utterances through a BiGRU, and combine the output and the input to form the memory bank: $\mathbf{M}_t = \{\overleftrightarrow{\GRU}(\mathbf{u}_{t-K-1+k}) + \mathbf{u}_{t-K-1+k}\}_{k=1}^{K}$. For simplicity, we use $\overleftrightarrow{\GRU}$ to denote the sum of the hidden states in the two directions of the BiGRU.

\subsection{Attention GRU}

Generally, the query in a conversation depends on the context it follows. Thus, it is crucial to weight and summarize the context to refine the representation of the query. This process is usually realized by an attention layer~\cite{DBLP:conf/emnlp/HazarikaPMCZ18,DBLP:conf/naacl/HazarikaPZCMZ18}, which allows the query to interact with the memory bank and produce a contextual vector $\mathbf{c}_t$.
\\[1ex]
\noindent\textbf{Context Weighting.} The attention layer should be able to retrieve relevant context from the memory bank to help predict the emotion expressed by the query. To weight the importance of each memory to the query, we adopt the dot-product attention with a $\softmax$ normalization. As a result, at $t$-step, the weight of the $k$th memory $\mathbf{M}_{t,k}$ will be:
\begin{align}
    a_k = \frac{\exp(\mathbf{q}_t^\top\mathbf{M}_{t,k})}{\sum_{k'=1}^K \exp(\mathbf{q}_t^\top\mathbf{M}_{t,k'})}.
\end{align}
\\[1ex]
\noindent\textbf{Context Summarizing.}
Conventionally, the contextual vector $\mathbf{c}_t$ can be produced by Soft Attention as a weighted sum of the memories, i.e., $\mathbf{c}_t = \sum_{k=1}^{K}a_k\mathbf{M}_{t,k}$. This method is efficient for computing, but just as the bag-of-words in word representation area~\cite{DBLP:conf/nips/MikolovSCCD13,DBLP:conf/emnlp/LingTAFDBTL15} it loses both the positional and ordering information of the memories. Thus, we propose an Attention GRU (AGRU) which uses the attention weight of the query to the memories to update the internal state $\tilde{\mathbf{h}}_t$ of a normal GRU. As a result, the output of the AGRU is:
\begin{align}
    \mathbf{h}_k = a_k\circ\tilde{\mathbf{h}}_k + (1 - a_k)\circ\mathbf{h}_{k-1}.
\end{align}
The GRU is advantageous for retaining the positional and ordering information of the memories and the attention weight controls the amount of information to be passed to the next step. We take the final hidden state of the AGRU as the contextual vector, i.e., $\mathbf{c}_t=\mathbf{h}_K$, then the refined query representation, i.e. the output, will be: 
\begin{align}
    \mathbf{o}_t = \mathbf{q}_t + \mathbf{c}_t.
\end{align}

Furthermore, considering the tendency of RNNs to better represent recent inputs~\cite{DBLP:journals/corr/BahdanauCB14}, the contextual vector from AGRU also tends to carry more information of the most recent memories. Accordingly, a backward AGRU can better represent memories distant from the query. Therefore, we promote AGRU to its bidirectional variant, i.e., BiAGRU, so as to make a balance between the information from recent memories and that from distant memories. We believe the BiAGRU is capable of capturing more comprehensive context from the memory bank, especially for long conversations. As a result, the contextual vectors produced by a BiAGRU are expressed as:
\begin{align}
    \mathbf{c}_t^f &= \overrightarrow{\AGRU}(\mathbf{M}_{t,K}, a_{K}, \overrightarrow{\mathbf{h}}_{K-1}), \\
    \mathbf{c}_t^b &= \overleftarrow{\AGRU}(\mathbf{M}_{t,1}, a_{1}, \overleftarrow{\mathbf{h}}_{2}),
\end{align}
which are used to refine the query representation similarly.

\subsection{Classifier}

The refined representation of the query from the AGRU is used for prediction by a $\softmax$ layer:
\begin{align}
    \hat{\mathbf{y}}_t = \softmax(\mathbf{W}_o\mathbf{o}_t + \mathbf{b}_o),
\end{align}
where $\mathbf{W}_o\in \mathbb{R}^{d_1\times|\mathcal{E}|}$, $\mathbf{b}_o\in \mathbb{R}^{|\mathcal{E}|}$, and $|\mathcal{E}|$ is the number of emotion classes.

We train the AGHMN model by a cross-entropy loss function, and the average loss of the whole training set can be computed as:
\begin{align}
\mathcal{L} &= \frac{1}{\sum_{l=1}^{L}T_l}\sum_{t=1}^{T_l}
\sum_{e=1}^{|\mathcal{E}|}\mathbf{y}_t^e\log(\hat{\mathbf{y}}_t^e),
\end{align}
where $T_l$ is the number of utterances in the $l$th conversation, and $L$ is the total number of conversations in the training set. $\mathbf{y}_t$ denotes the one-hot vector of the target emotion labels, and $\mathbf{y}_t^e$ and $\hat{\mathbf{y}}_t^e$ are the elements of $\mathbf{y}_t$ and $\hat{\mathbf{y}}_t$ for the emotion class $e$, respectively.

\begin{table}[t]
\small
\centering
\begin{threeparttable}
\begin{tabular}{c|c|c|c|c}
\toprule
\multirow{2}{*}{Dataset}
& \multirow{2}{*}{Fold}
& \multicolumn{1}{c|}{\footnotesize{No. of}}
& \multicolumn{1}{c|}{\footnotesize{No. of}}
& \multicolumn{1}{c}{\footnotesize{Avg. length}} \\
& & \footnotesize{Utt} & \footnotesize{Conv} & \footnotesize{of Conv} \\
\hline
\hline
\multirow{2}{*}{IEMOCAP}
& train/val & {5810} & 120 & 48.4 \\
& test & {1623} & 31& 52.4 \\
\multirow{2}{*}{MELD}
& train/val & {11098} & {1153}& 9.6 \\
& test & {2610} & 280& 9.3 \\
\bottomrule
\end{tabular}
\end{threeparttable}
\caption{\label{table:datasets}Summary of the two conversation emotion datasets: IEMOCAP, and MELD.}
\end{table}

\begin{table*}[t!]
\small
\centering
\begin{threeparttable}
\resizebox{0.95\textwidth}{!}{
\begin{tabular}{l|cc|cc|cc|cc|cc|cc|ccc}
\toprule
\multirow{3}{*}{Model} 
& \multicolumn{15}{c}{\bf IEMOCAP: Emotion Classes} \\
\cline{2-16}
& \multicolumn{2}{c|}{\it happy}
& \multicolumn{2}{c|}{\it sad}
& \multicolumn{2}{c|}{\it neutral}
& \multicolumn{2}{c|}{\it angry}
& \multicolumn{2}{c|}{\it excited}
& \multicolumn{2}{c|}{\it frustrated}
& \multicolumn{3}{c}{\bf Avg.} \\
\cline{2-16}
& Acc & F1 & Acc & F1 & Acc & F1 & Acc & F1 & Acc & F1 & Acc & F1 & Acc & F1 & mF1 \\
\hline
\hline
memnet$^\triangle$ & 25.7& 33.5& 55.5& 61.7& 58.1& 52.8& 59.3& 55.3& 51.5& 58.3& 67.2& 59.0& 55.7& 55.1 & 53.4 \\
CMN$^\triangle$ & 25.0& 30.3& 55.9& 62.4& 52.8& 52.3& 61.7& 59.8& 55.5& 60.2& \bf71.1& 60.6& 56.5& 56.1& 54.3 \\
ICON$^\diamondsuit$ & -& -& -& -& -& -& -& -& -& -& -& -& 58.3& 57.9& - \\
DialogueRNN$^\triangle$ & 31.2& 33.8& 66.1& 69.8& 63.0& 57.7& 61.7& \bf62.5& 61.5& 64.4& 59.5& 59.4& 59.3& 59.8& 57.9 \\
\hline
scLSTM & 37.5& 43.4& 67.7& 69.8& \bf64.2& 55.8& 61.9& 61.8& 51.8& 59.3& 61.3& 60.2& 59.2& 59.1& 58.4 \\
DialogueRNN & 33.5& 35.4& \bf69.0& 68.7& 54.1& 54.7& \bf67.1& 61.1& 55.9& 60.4& 62.9& 60.3& 58.3& 58.1& 56.7 \\
\hline
UniF-AGRU & 42.7& 51.1& 63.4& 68.0& 61.3& 57.4& 61.9& 61.8& 67.5& \bf70.5& 64.1& 60.5& 61.9& 61.8& 61.5 \\
UniF-BiAGRU & 49.7& 50.6& 64.7& 69.9& 60.3& 59.0& 55.1& 60.5& 67.4& 69.6& 68.6& 62.1& 62.8& 62.7& 61.9 \\
BiF-AGRU & 48.3& 52.1& 68.3& \bf73.3& 61.6& 58.4& 57.5& 61.9& \bf68.1& 69.7& 67.1& \bf62.3& \bf63.5& \bf63.5& \bf63.0 \\
BiF-BiAGRU & \bf62.1& \bf54.5& 66.6& 72.7& 63.9& \bf59.4& 58.4& 61.0& 58.5& 66.6& 64.8& 61.6& 62.8& 63.0& 62.6 \\
\bottomrule
\end{tabular}
}
\end{threeparttable}
\caption{\label{table:result_IEMOCAP}Performance of AGHMN models on IEMOCAP.}
\end{table*}

\begin{table*}[t!]
\centering
\begin{threeparttable}
\resizebox{0.95\textwidth}{!}{
\begin{tabular}{l|cc|cc|cc|cc|cc|cc|cc|ccc}
\toprule
\multirow{3}{*}{Model} 
& \multicolumn{17}{c}{\bf MELD: Emotion Classes} \\
\cline{2-18}
& \multicolumn{2}{c|}{\it neutral}
& \multicolumn{2}{c|}{\it surprise}
& \multicolumn{2}{c|}{\it fear}
& \multicolumn{2}{c|}{\it sadness}
& \multicolumn{2}{c|}{\it joy}
& \multicolumn{2}{c|}{\it disgust}
& \multicolumn{2}{c|}{\it anger}
& \multicolumn{3}{c}{\bf Avg.} \\
\cline{2-18}
& Acc & F1 & Acc & F1 & Acc & F1 & Acc & F1 & Acc & F1 & Acc & F1& Acc & F1 & Acc & F1 & mF1 \\
\hline
\hline
CNN$^\dagger$ & -& 74.9& -& 45.5& -& 3.7& -& 21.1& -& 49.4& -& 8.2& -& 34.5 &- & 55.0& 33.9 \\
scLSTM & 78.4& 73.8& 46.8& 47.7& 3.8& 5.4& 22.4& 25.1& 51.6& 51.3& 4.3& 5.2& 36.7& 38.4& 57.5& 55.9& 35.3 \\
DialogueRNN & 72.1&	73.5& \bf54.4&	49.4& 1.6&	1.2& \bf23.9& 23.8& 52.0&	50.7& 1.5&	1.7& \bf41.9& \bf41.5& 56.1& 55.9& 34.5 \\
\hline
UniF-AGRU & 80.3& 75.1& 53.7& 49.1& \bf9.8& 10.6& 19.7& 25.5& 50.5& 51.1& \bf14.0& 16.4& 33.9& 38.2& 58.8& 57.0& 38.0 \\
UniF-BiAGRU & \bf83.4& \bf76.4& 49.1& \bf49.7& 9.2& \bf11.5& 21.6& \bf27.0& 52.4& \bf52.4& 12.2& 14.0& 34.9& 39.4& \bf60.3& \bf58.1& \bf38.6 \\
BiF-AGRU & 81.6& 75.5& 50.5& 49.0& 8.0& 10.4& 22.3& 26.9& 51.7& 51.7& 13.7& \bf17.2& 34.5& 38.6& 59.5& 57.5& 38.5 \\
BiF-BiAGRU & 80.7& 75.4& 50.7& 47.9& 7.2& 10.0& 20.9& 26.3& \bf53.9& 52.1& 10.4& 12.8& 33.9& 38.1& 59.2& 57.1& 37.5 \\
\bottomrule
\end{tabular}
}
\end{threeparttable}
\caption{\label{table:result_MELD}Performance of AGHMN models on MELD.}
\end{table*}

\section{Experimental Setup}

In this section, we will present the details of our experimental setup, including datasets, compared methods, implementation, and training.
\\[1ex]
\noindent\textbf{Datasets.}
We train and test our model on two conversation emotion datasets, namely, IEMOCAP~\cite{DBLP:journals/lre/BussoBLKMKCLN08}, and MELD~\cite{DBLP:conf/acl/PoriaHMNCM19}.

$\bullet$~\textbf{IEMOCAP\footnote{\url{https://sail.usc.edu/iemocap/}}}:
The IEMOCAP dataset contains the acts of 10 speakers in a dyadic conversation fashion, providing text, audio, and video features. We follow the previous work~\cite{DBLP:conf/emnlp/HazarikaPMCZ18} to use the first four sessions of transcripts as the training set, and the last one as the testing set. The validation set is extracted from the randomly-shuffled training set with the ratio of 80:20. Also, we focus on recognizing six emotion classes, namely, \textit{happy}, \textit{sad}, \textit{neutral}, \textit{angry}, \textit{excited}, and \textit{frustrated}.

$\bullet$~\textbf{MELD\footnote{\url{https://github.com/SenticNet/MELD}}}:
The MELD dataset~\cite{DBLP:conf/acl/PoriaHMNCM19} is an extended version of the EmotionLines dataset~\cite{DBLP:conf/lrec/HsuCKHK18}. The data comes from the Friends TV series with multiple speakers involved in the conversations. It is split into training, validation, and testing sets with 1039, 114, and 280 conversations, respectively. Each utterance has been labelled by one of the seven emotion types, namely, \textit{anger}, \textit{disgust}, \textit{sadness}, \textit{joy}, \textit{neutral}, \textit{surprise} and \textit{fear}. 
\\[1ex]
\noindent\textbf{Compared Methods.}
With the different combination of memory banks and AGRUs, we consider four variants of AGHMN\footnote{\url{https://github.com/wxjiao/AGHMN}} for experiments: UniF-AGRU, UniF-BiAGRU, BiF-AGRU, and BiF-BiAGRU.
These variants are compared to the following baselines:

$\bullet$~\textbf{scLSTM}~\cite{DBLP:conf/acl/PoriaCHMZM17} is the unidirectional variant that classifies utterances using historical utterance as context realized by a LSTM.

$\bullet$~\textbf{CMN}~\cite{DBLP:conf/naacl/HazarikaPZCMZ18} models separate contexts for both speaker and listener to an utterance. These contexts are stored as memories to aid the prediction of an incoming utterance.

$\bullet$~\textbf{DialogueRNN}~\cite{DBLP:conf/aaai/MajumderPHMGC19} is the unidirectional variant with an attention layer that keeps track of the individual party states throughout the conversation and uses this information for emotion classification.

$\bullet$~\textbf{ICON}~\cite{DBLP:conf/emnlp/HazarikaPMCZ18} incorporates the self- and inter-speaker emotional influences into global memories to help predict the emotional orientation of utterances. It is a unidirectional model with only historical context.

For IEMOCAP, we refer to the results of \textbf{memnet}~\cite{DBLP:conf/nips/SukhbaatarSWF15}, CMN, DialogueRNN from $^\triangle$~\cite{DBLP:conf/aaai/MajumderPHMGC19}, and that of ICON from $^\diamondsuit$~\cite{DBLP:conf/emnlp/HazarikaPMCZ18}. For MELD, we refer to the results of \textbf{CNN} from $^\dagger$~\cite{DBLP:journals/access/PoriaMMH19}. We re-run scLSTM and DialogueRNN for both datasets.
CMN and ICON cannot be adapted for MELD because they are customized for dyadic conversations and may encounter scalability issue for multiparty conversation datasets~\cite{DBLP:journals/access/PoriaMMH19}. 
\\[1ex]
\noindent\textbf{Implementation.}
We implement scLSTM and our proposed AGHMN models from scratch on the Pytorch\footnote{\url{https://pytorch.org/}} framework. For the extraction of textual feature, we follow~\cite{DBLP:conf/emnlp/HazarikaPMCZ18} to adopt a 1-D CNN with filters of 3, 4, and 5 each with 64 feature maps. The convolution result of each filter is fed to a max-over-time pooling layer. The pooling results are concatenated and transformed to the utterance embeddings via a $\relu$ layer. The hidden size of the LSTM is 100. As for our AGHMN models, the hidden sizes of GRUs and AGRUs are also 100. By default, the context-window size $K$ for the memory bank is 40 for IEMOCAP and 10 for MELD, which are around the average conversation lengths of each dataset, respectively. Besides, the implementation of DialogueRNN comes from the open-source codes\footnote{\url{https://github.com/SenticNet/conv-emotion/tree/master/DialogueRNN}} provided by the authors of DialogueRNN.
\\[1ex]
\noindent\textbf{Training.}
We choose Adam~\cite{DBLP:journals/corr/KingmaB14} optimizer with an initial learning rate $lr = 5\times 10^{-4}$. To regulate the models, we clip the gradients of model parameters with a max norm of 5 and apply dropout with a drop rate of 0.3. We monitor the macro-averaged F1-score (mF1) of the validation sets during training and decay the learning rate by 0.95 once the mF1 stops increasing. The training process is terminated by early stopping with a patience of 10.

\section{Results}

Table~\ref{table:result_IEMOCAP} and Table~\ref{table:result_MELD} present the results on IEMOCAP and MELD testing sets, respectively. For both datasets, we report the accuracy and F1-score~\cite{DBLP:conf/acl/TongZJM17} for each emotion class and evaluate the overall classification performance using their weighted averages of all emotion classes. We also report the macro-average of F1-score (mF1) to reflect the model performance on minority emotion classes, since the weighted-average is compromised by the majority classes. Each result provided by us in the tables is the average value of 10 times repeated experiments.

In Table~\ref{table:result_IEMOCAP}, all of our AGHMN models perform better than the compared models. BiF-AGRU attains the best overall performance with significant improvement over the strongest baseline DialogueRNN$^\triangle$ (+4.2\% Acc, +3.7\% F1, +5.1\% mF1). For each emotion, our ADHMN models achieve at least competitive performance as DialogueRNN$^\triangle$. In particular, our models attain very large improvement on \textit{happy} (at least +11.5\% Acc, +17.3\% F1), which is the emotion with the least utterances. This demonstrates the capability of our models in recognizing minority emotion classes.

In Table~\ref{table:result_MELD}, all the AGHMN models also outperform the compared methods significantly. But this time, UniF-BiAGRU becomes the best one (+2.8\% Acc, +2.2\% F1, +3.3\% mF1).
Our models perform the best on most emotion classes, especially the two minority classes \textit{fear} and \textit{disgust}, though this is accompanied by the performance degradation on \textit{anger}. But referring to the mF1 value, it is safe to say that our models produce much more balanced results. 
\\[1ex]
\noindent\textbf{Baseline Methods.}
The bcLSTM model implemented by us is very strong, performing better than all the other baselines on both datasets except DialogueRNN$^\triangle$ on IEMOCAP. However, the results of DialogueRNN run by us are slightly worse than DialogueRNN$^\triangle$. This may be because we follow the default settings of the provided codes, which are customized for BiDialogueRNN. No matter what, our AGHMN models outperform them on the two datasets, suggesting the efficacy of our context-modeling scheme.
\\[1ex]
\noindent\textbf{AGHMN variants.} 
UniF-AGRU performs the worst among all the four variants on both datasets. Consistently, UniF-BiAGRU and BiF-AGRU outperform UniF-AGRU, which demonstrates the superiority of BiAGRU over AGRU and BiF over UniF, respectively.
However, BiF-BiAGRU does not attain the best performance, which we speculate that the model becomes too deep to learn from the two datasets. In fact, the performance difference between the variants on MELD is limited. This is mainly because the conversations in MELD contain much fewer turns than that in IEMOCAP, making it less sensitive to different modules.

\begin{table}[t!]
\small
\centering
\begin{threeparttable}
\begin{tabular}{l|ccc|ccc}
\toprule
\multirow{2}{*}{Reader} 
& \multicolumn{3}{c|}{\bf UniF-BiAGRU}
& \multicolumn{3}{c}{\bf BiF-AGRU} \\
\cline{2-7}
& Acc & F1 & mF1 & Acc & F1 & mF1 \\
\hline
\hline
CNN & 59.8&	59.9& 59.4& 60.2& 60.2& 59.9 \\
CNN$_{soft}$ & 58.8& 58.7& 58.1& 59.4& 59.3& 58.8 \\
BiLSTM & 62.0& 62.0& 61.2& 62.8& 62.7& 61.9 \\
BiGRU & \bf62.8& \bf62.7& \bf61.9& \bf63.5& \bf63.5& \bf63.0 \\
\bottomrule
\end{tabular}
\end{threeparttable}
\caption{\label{table:result_ReaderCompare}Testing results on IEMOCAP, with CNN, BiLSTM, and BiGRU as the utterance reader, respectively.}
\end{table}

\begin{table}[t!]
\fontsize{10}{11}\selectfont
\centering
\begin{threeparttable}
\resizebox{0.95\columnwidth}{!}{
\begin{tabular}{l|ccc|ccc}
\toprule
\multirow{2}{*}{Mem-Attn}
& \multicolumn{3}{c|}{\bf IEMOCAP}
& \multicolumn{3}{c}{\bf MELD} \\
\cline{2-7}
& Acc & F1 & mF1 & Acc & F1 & mF1 \\
\hline
\hline
UniF-Soft & 60.7& 60.5& 59.5& 58.0& 56.5& 37.7 \\
BiF-Soft & 62.3& 62.2& 61.8& 58.6& 56.9& 37.5 \\
\hline
UniF-AGRU & 61.9& 61.8& 61.5& 58.8& 57.0& 38.0 \\
UniF-BiAGRU & \bf62.8& \bf62.7& \bf61.9& \bf60.3& \bf58.1& \bf38.6 \\
\bottomrule
\end{tabular}
}
\end{threeparttable}
\caption{\label{table:result_AttCompare}Testing results on IEMOCAP and MELD, with different choices of attention and memory bank.}
\end{table}

\begin{figure}[t]
    \centering
    \includegraphics[width=0.95\columnwidth]{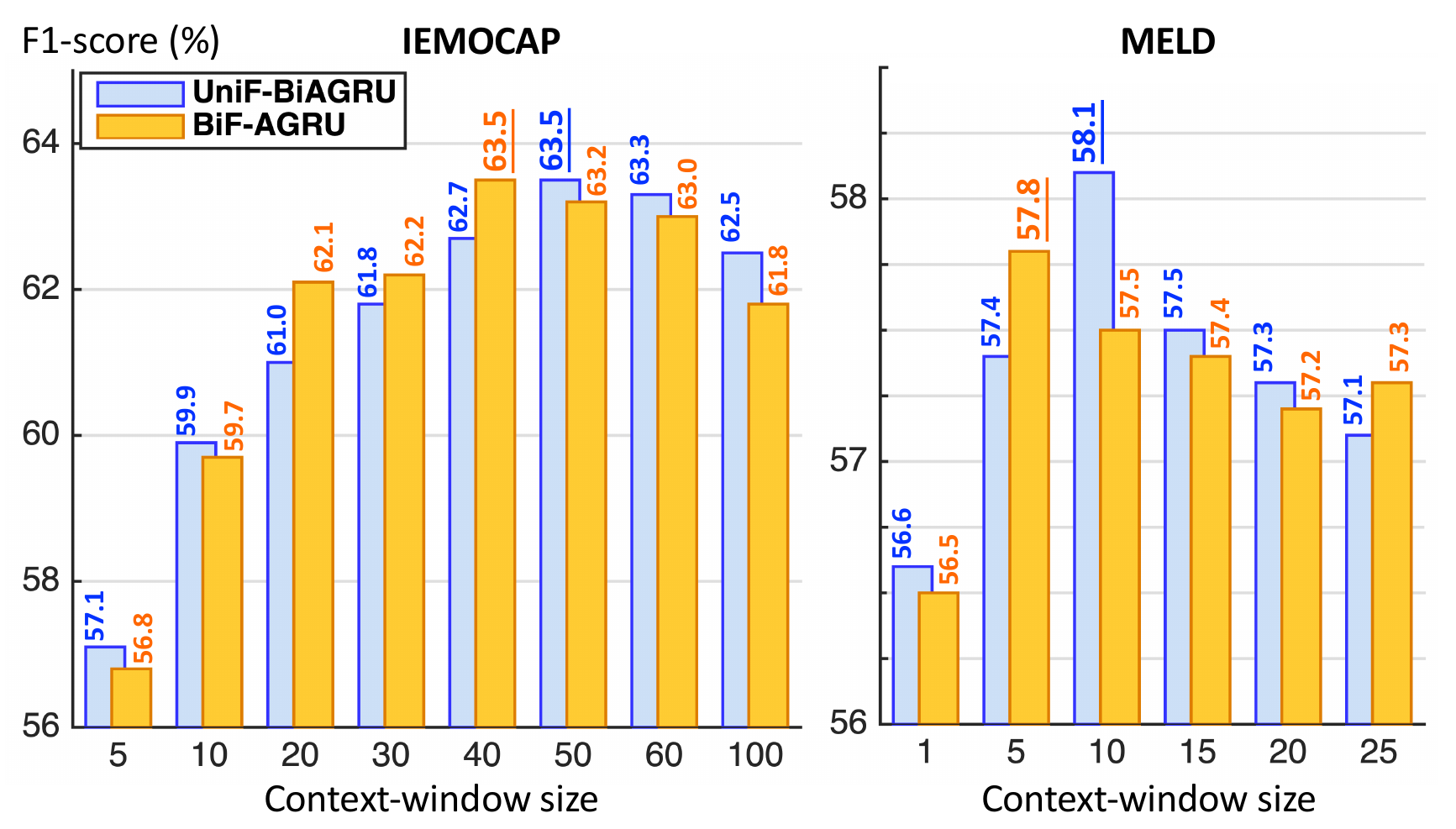}
    \caption{Trends in the performance of UniF-BiAGRU and BiF-AGRU with varying context-window size $K$.}
    \label{fig:context_window}
\end{figure}

\begin{figure*}[t!]
    \centering
    \includegraphics[width=0.95\textwidth]{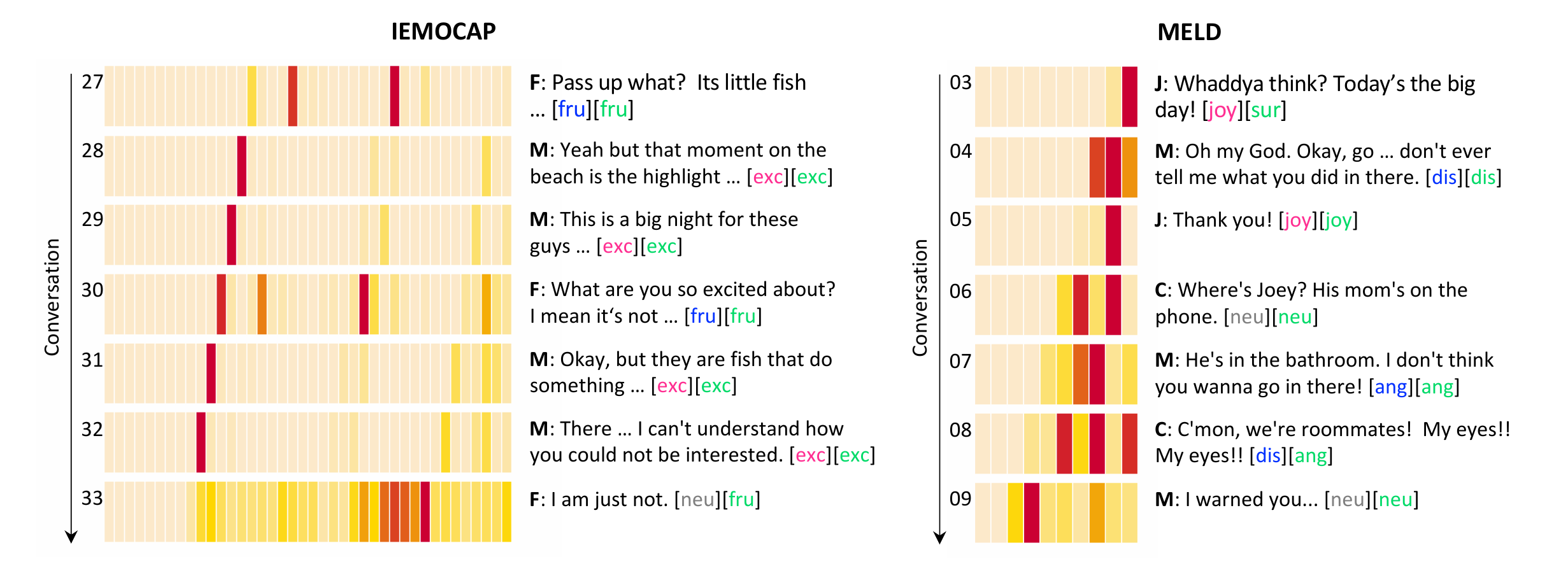}
    \caption{Evolution of memory selection as a conversation develops. The attention weights come from BiF-AGRU for both IEMOCAP and MELD. Each utterance is tagged with two labels, the first is ground truth and the second is the prediction by BiF-AGRU. IEMOCAP: M: Male, F: Female; MELD: J: Joey, M: Monica, C: Chandler.}
    \label{fig:att_visual1}
\end{figure*}

\begin{figure}[t]
    \centering
    \includegraphics[width=0.95\columnwidth]{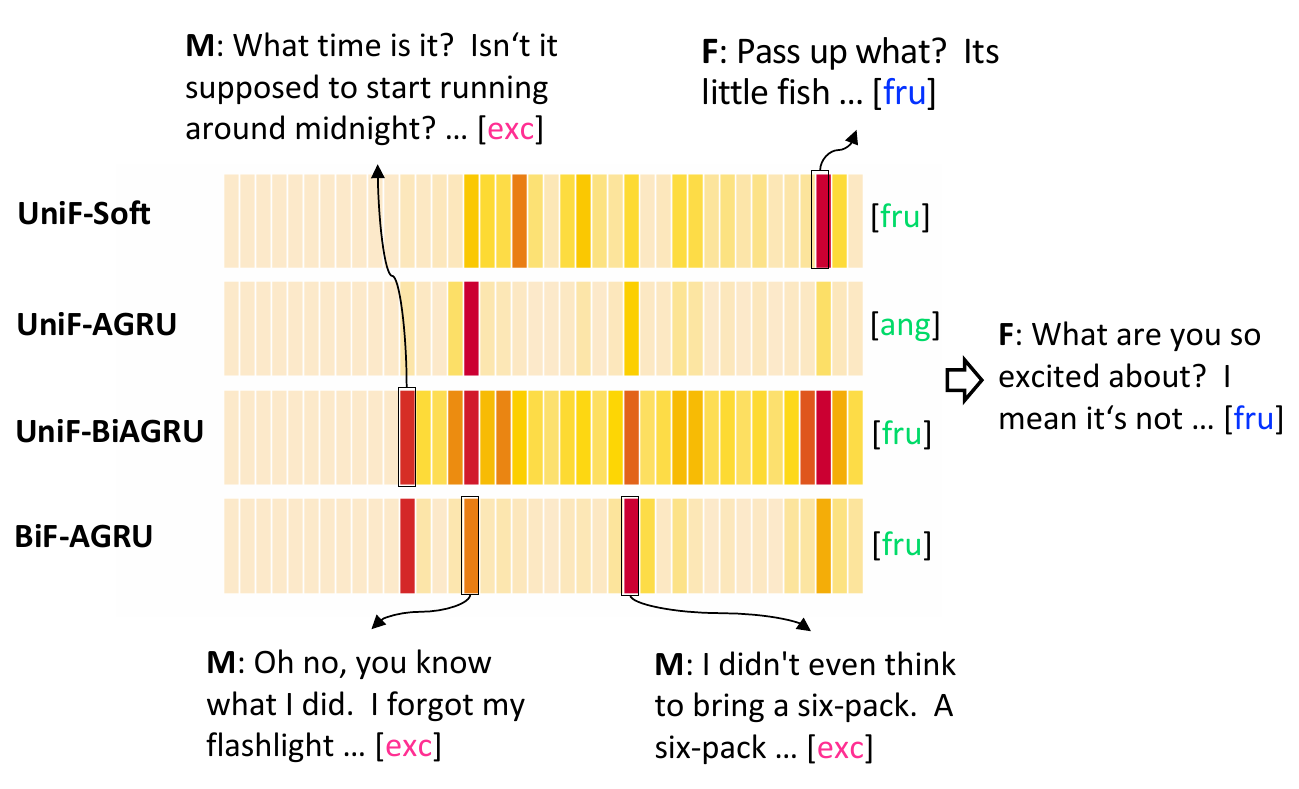}
    \caption{Comparison of memory selection between UniF-Soft and UniF-AGRU, UniF-BiAGRU, and BiF-AGRU.}
    \label{fig:att_visual2}
\end{figure}

\subsection{Model Analysis}
\label{ssec:model_analysis}

\noindent\textbf{Utterance Readers.}
We argue that the BiGRU is a better reader for utterances. Here, we test UniF-BiAGRU and BiF-AGRU on IEMOCAP with the utterance reader replaced by a 1-D CNN and a bidirectional LSTM (BiLSTM), respectively. The results are reported in Table~\ref{table:result_ReaderCompare}, where we can easily conclude that the RNN variants (BiLSTM, and BiGRU) surpass 1-D CNN with significant margins. For BiLSTM and BiGRU, the latter one performs better because GRUs are usually more powerful than LSTMs on small datasets. These results indicate that RNNs are more compatible in this Hierarchical Memory Network. Moreover, as BiGRU, BiLSTM attains better performance in BiF-AGRU than in UniF-BiAGRU, suggesting the advantage of BiF. In addition, we also conduct experiments for the CNN reader with Soft Attention, named as CNN$_{soft}$, which encounters the degradation of performance. It demonstrates that the improvement of our models is not achieved by only the BiGRU reader. 
\\[1ex]
\noindent\textbf{Attention Choices \& Memory Banks.}
We investigate the advantage of AGRU over Soft Attention here. As presented in Table~\ref{table:result_AttCompare}, with the UniF memory bank, AGRU attains better results than Soft Attention on both datasets. The bidirectional variant of AGRU extends the advantage even further. It is noteworthy that we also include the results of Soft Attention with the BiF memory, which are considerably better than that with UniF. This further verifies that BiF can produce better memory representation.
\\[1ex]
\noindent\textbf{Context-Window Size.}
We plot the performance trend of UniF-BiAGRU and BiF-AGRU on both datasets when varying the context-window size $K$ for building the memory bank. On both datasets, the two models follow similar trends such that the performance increases at first and falls then with the increase of $K$. On IEMOCAP, the best results are obtained at $K=40$ for BiF-AGRU and $K=50$ for UniF-BiAGRU, which align with the average length of conversations in the dataset (see Table~\ref{table:datasets}). On MELD, the best values of $K$ are 5 and 10, respectively, which are much lower than that on IEMOCAP. We speculate that it is because the data of MELD comes from the Friends TV sitcom with more rapid emotion fluctuation. Therefore, a longer context may result in model confusion. In contrast, the emotion state in IEMOCAP evolves much more gently.

\subsection{Case Study}
\label{ssec:case_study}

\noindent\textbf{Attention Evolution.}
We find that the selection of memory differs between speakers as a conversation develops, though we do not distinguish the speakers explicitly in our model. In Fig~\ref{fig:att_visual1}, we visualize the attention weights of BiF-AGRU tested on a conversation fragment of IEMOCAP and MELD, respectively. For IEMOCAP, the Male is \textit{excited} from the beginning so that the four utterances presented here pay the most attention to the first utterance of the conversation. In contrast, the attention of the Female is distributed over several historical utterances, including the first one and some intermediate ones (see Fig~\ref{fig:att_visual2}) that make her \textit{frustrated}. As for MELD, Joey is joyful all the time and his attention is paid to his last \textit{joyful} utterance. Monica pays her most attention to the conversation between her and Joey, providing clues for Chandler. Chandler focuses on Joey and where Joey is.
\\[1ex]
\noindent\textbf{Attention Comparison.}
The selection of memory also varies between different attention mechanisms. In Fig~\ref{fig:att_visual2}, given a query utterance expressed by the Female with \textit{frustrated} emotion, Soft Attention focuses on the utterance that also expresses \textit{frustrated}. AGRU pays most attention to one utterance that could be the reason for the Female's \textit{frustrated} emotion, but it classifies the query emotion as \textit{angry} in this example. BiAGRU can sense both kinds of clues, providing more comprehensive memory. BiF improves the memory representation and helps AGRU to extract the memory as comprehensively as UniF-BiAGRU.
\\[1ex]
\noindent\textbf{Error Analysis.}
In Fig~\ref{fig:att_visual1}, the 33rd utterance of IEMOCAP is recognized as \textit{frustrated} not \textit{neutral}. We argue that the original annotation might not be accurate. Given several turns of \textit{frustrated} and the latest response (34th) from the Male, the Female could express \textit{frustrated} again. Still, we cannot deal with minority classes very well on MELD, such as \textit{disgust}. With more data or multimodal features to disambiguate with other emotions, this issue might be better resolved.

\section{Conclusions}

We propose an Attention Gated Hierarchical Memory Network (AGHMN) for Real-Time Emotion Recognition. Firstly, the proposed Hierarchical Memory Network improves the quality of utterance features and memories. Then, the proposed Attention GRU summarizes better contextual information than the commonly used Soft Attention. We conduct extensive experiments on two emotion conversation datasets, and our models outperform the state-of-the-art approaches with significant margins. Lastly, ablation studies and attention visualization demonstrate the efficacy of each component of our AGHMN models.

\section{ Acknowledgments}
This work is supported by the Research Grants
Council of the Hong Kong Special Administrative Region, China (No. CUHK 14208815 and
No. CUHK 14210717 of the General Research
Fund). We thank the three anonymous
reviewers for their insightful suggestions on various
aspects of this work.

\bibliographystyle{aaai}
\bibliography{aaai}
\end{document}